\pdfoutput=1
\documentclass[11pt]{article}
\usepackage[final]{acl}
\usepackage{times}
\usepackage{latexsym}
\usepackage[T1]{fontenc}
\usepackage[utf8]{inputenc}
\usepackage{microtype}
\usepackage{inconsolata}
\usepackage{graphicx}
\usepackage{hyperref}
\usepackage{amsmath}
\usepackage{amssymb}
\usepackage{mathtools}
\usepackage{amsthm}
\usepackage{bm}
\usepackage{thmtools}
\usepackage{thm-restate}
\usepackage{multirow,adjustbox,makecell,arydshln} 
\usepackage{diagbox}
\usepackage{longtable}
\usepackage{color, colortbl}
\usepackage{booktabs}
\usepackage{xcolor}

\usepackage{pifont}
\newcommand{\cmark}{\ding{51}} % ✓
\newcommand{\xmark}{\ding{55}} % ✗
\theoremstyle{plain}

\newtheorem*{theoremr*}{Theorem~\ref{th1}}
\newtheorem*{theoremr2*}{Theorem~\ref{th2}}
\newtheorem*{corr*}{Corollary~\ref{cor1}}
\newtheorem*{propositionr*}{Proposition~\ref{prop1}}
\newtheorem*{propositionr2*}{Proposition~\ref{p2}}
\theoremstyle{definition}

\theoremstyle{remark}

\usepackage{amsfonts}
\usepackage{enumitem} 
\usepackage{tcolorbox} % for examples in text

\definecolor{aquamarine}{rgb}{0.5, 1.0, 0.83}
\definecolor{ao}{rgb}{0.0, 0.5, 0.0}

\usepackage{listings}

\title{DTECT: Dynamic Topic Explorer \& Context Tracker}

\author{Suman Adhya \and Debarshi Kumar Sanyal \\
    Indian Association for the Cultivation of Science\\
         \texttt{\href{mailto:adhyasuman30@gmail.com}{adhyasuman30@gmail.com}, \href{mailto:debarshi.sanyal@iacs.res.in}{debarshi.sanyal@iacs.res.in}}
}

\begin{document}
\maketitle
\begin{abstract}
The explosive growth of textual data over time presents a significant challenge in uncovering evolving themes and trends. Existing dynamic topic modeling techniques, while powerful, often exist in fragmented pipelines that lack robust support for interpretation and user-friendly exploration. We introduce \textbf{DTECT} (\textbf{D}ynamic \textbf{T}opic \textbf{E}xplorer \& \textbf{C}ontext \textbf{T}racker), an end-to-end system that bridges the gap between raw textual data and meaningful temporal insights. DTECT provides a unified workflow that supports data preprocessing, multiple model architectures, and dedicated evaluation metrics to analyze the topic quality of temporal topic models. It significantly enhances interpretability by introducing LLM-driven automatic topic labeling, trend analysis via temporally salient words, interactive visualizations with document-level summarization, and a natural language chat interface for intuitive data querying. By integrating these features into a single, cohesive platform, DTECT empowers users to more effectively track and understand thematic dynamics. DTECT is open-source and available at \url{https://github.com/AdhyaSuman/DTECT}.
\end{abstract}

\section{Introduction}
Understanding how topics evolve over time is crucial for interpreting large, temporally structured text corpora such as news archives, academic literature, and social media. The Dynamic Topic Model (DTM) \cite{blei2006dynamic} extends Latent Dirichlet Allocation (LDA) \cite{blei2003latent} to model topic trajectories over time.

Subsequent advancements include the Dynamic Embedded Topic Model (DETM) \cite{dieng2019dynamic}, the Chain-Free Dynamic Topic Model (CFDTM) \cite{wu2024dynamic}, and evaluation metrics specific for dynamic topic models by \citet{charu2024evaluating}.

Several toolkits such as \textsc{Gensim} \cite{rehurek2010gensim}, \textsc{OCTIS} \cite{terragni2021octis}, and \textsc{TopMost} \cite{wu2024topmost} support topic modeling workflows.

Despite these advances, a unified end-to-end system for dynamic topic modeling is still missing. Semantic interpretability remains a key challenge \cite{doogan2021interpretability}, as topics are typically unlabeled distributions over words. This makes it tedious to inspect and track topic shifts across time.

To address these limitations, we present \textbf{DTECT} (\textbf{D}ynamic \textbf{T}opic \textbf{E}xplorer and \textbf{C}ontext \textbf{T}racker), a unified, interactive system for dynamic topic modeling and interpretation:

\begin{tcolorbox}[
    title=\centering \textbf{Contributions},
    colback=blue!3,
    colframe=blue!60,
    coltitle=white,
    fonttitle=\bfseries,
    rounded corners,
    left=-2mm, right=1mm, top=1mm, bottom=1mm,
    boxrule=0.6pt
]
\begin{itemize}
    \item[\textcolor{blue}{\ding{72}}] An end-to-end pipeline for dynamic topic modeling, covering preprocessing, modeling (DTM, DETM, CFDTM), and evaluation (TTC, TTS, TTQ).
    
    \item[\textcolor{blue}{\ding{72}}] LLM-enhanced topic labeling for interpreting evolving topics across time.
    
    \item[\textcolor{blue}{\ding{72}}] A scoring-based method to surface temporally informative words and visualize their trajectories.
    
    \item[\textcolor{blue}{\ding{72}}] Context-specific document retrieval via keyword-timestamp queries, with document highlighting.
    
    \item[\textcolor{blue}{\ding{72}}] LLM-based summarization of retrieved documents to explain temporal trends.
    
    \item[\textcolor{blue}{\ding{72}}] A conversational interface for grounded, follow-up exploration based on retrieved evidence.
\end{itemize}
\end{tcolorbox}

\section{System Overview}
\begin{figure}[ht]
    \centering
    \includegraphics[width=\linewidth]{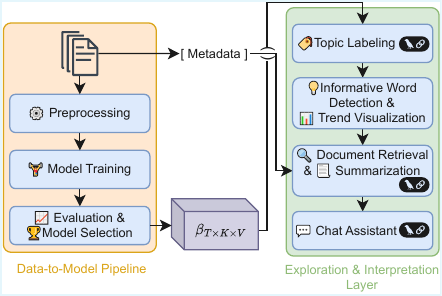}
    \caption{System architecture and workflow of DTECT.}
    \label{fig:workflow}
\end{figure}

As illustrated in Figure~\ref{fig:workflow}, DTECT comprises two main components: (1) \textit{Data-to-Model Pipeline} and (2) \textit{Exploration \& Interpretation Layer}.

\paragraph{Data-to-Model Pipeline.} This module covers: (a) \textit{Preprocessing} of timestamped documents via standard NLP methods; (b) \textit{Model Training} using three dynamic topic models, and (c) \textit{Evaluation \& Selection} based on the temporal evaluation scores and runtime.

\paragraph{Exploration \& Interpretation Layer.} This layer supports intuitive understanding through: (a) \textit{Topic Labeling} using LLMs based on temporal keyword trajectories; (b) \textit{Informative Word Detection \& Trend Exploration}, which scores topic-specific words based on their temporal distributions and visualizes their evolution over time; (c) \textit{Document Retrieval \& Summarization}, allowing users to retrieve timestamp-specific documents for selected keywords and generate concise summaries to highlight key developments; (d) an interactive \textit{Chat Assistant} that supports natural language queries grounded in the generated summaries for deeper pattern exploration.

The next sections detail its two core components.

\section{Data-to-Model Pipeline}
DTECT provides a flexible pipeline to process timestamped corpora and fit dynamic topic models. This section describes the datasets, preprocessing steps, supported models, and evaluation methods.

\subsection{Datasets and Preprocessing}
We showcase DTECT’s capabilities across three corpora spanning academic, diplomatic, and legislative domains (Table~\ref{tab:datasets}). Each dataset is processed using a modular OCTIS-based pipeline, producing tokenized texts, vocabulary, BoW matrices, and timestamp indices. Figure~\ref{fig:code-preprocessing} illustrates how users can apply the preprocessing script to their own custom datasets and start exploring.

\begin{table}[ht]
    \centering
    \begin{adjustbox}{width=\linewidth}
    \begin{tabular}{lcccc}
        \toprule
        \textbf{Dataset} & \textbf{\#Docs} & \textbf{Avg Len} & \textbf{Std Len} & \textbf{\#Timestamps} \\
        \midrule
        ACL Anthology & 78,622 & 530.5 & 332.0 & 16 (2010--2025) \\
        UN Debates    & 7,507  & 10,847.2 & 4,671.3 & 46 (1970--2015) \\
        TCPD-IPD Finance & 20,361 & 858.9 & 690.3 & 21 (1999--2019) \\
        \bottomrule
    \end{tabular}
    \end{adjustbox}
    \caption{Dataset statistics: number of documents, average and standard deviation of document lengths (in words), and the number of temporal bins.}
    \label{tab:datasets}
\end{table}

\begin{figure}[ht]
    \vspace{-\intextsep}
    \begin{lstlisting}[language=Python]
from backend.datasets.preprocess import Preprocessor
from nltk.corpus import stopwords
import os

dataset_dir = '../data/Sample_data/'
stop_words = stopwords.words('english')

preprocessor = Preprocessor(
    docs_jsonl_path=dataset_dir + 'docs.jsonl',
    output_folder=os.path.join(dataset_dir, 'processed'),
    use_partition=False,
    min_count_bigram=5,
    threshold_bigram=20,
    remove_punctuation=True,
    lemmatize=True,
    stopword_list=stop_words,
    min_chars=3, 
    min_words_docs=3,
)
preprocessor.preprocess()
    \end{lstlisting}
    \caption{Example: DTECT preprocessing pipeline.}
    \label{fig:code-preprocessing}
\end{figure}

\begin{figure}[ht]
    \vspace{-\intextsep}
    \begin{lstlisting}[language=Python, captionpos=b]
from backend.datasets import dynamic_dataset
from backend.models.CFDTM.CFDTM import CFDTM
from backend.models.dynamic_trainer import DynamicTrainer
from backend.evaluation.eval import TopicQualityAssessor

# Load dataset
data = dynamic_dataset.DynamicDataset('../data/Sample_data/processed')

# Initialize model
model = CFDTM(
    vocab_size=data.vocab_size,
    num_times=data.num_times,
    num_topics=20,
    pretrained_WE=data.pretrained_WE,
    train_time_wordfreq=data.train_time_wordfreq
).to("cuda")

# Train model
trainer = DynamicTrainer(model, data)
top_words, _ = trainer.train()
top_words_list = [[topic.split() for topic in timestamp] for timestamp in top_words]
train_corpus = [doc.split() for doc in data.train_texts]

# Evaluation
assessor = TopicQualityAssessor(
    topics=top_words_list,
    train_texts=train_corpus,
    topn=10,
    coherence_type='c_npmi'
)
summary = assessor.get_dtq_summary()
    \end{lstlisting}
    \caption{Example training and evaluation pipeline.}
    \label{fig:code_training}
\end{figure}
    
\subsection{Supported Models}
DTECT supports three dynamic topic models: the classical DTM~\cite{blei2006dynamic}, which extends LDA with temporal dependencies; DETM~\cite{dieng2019dynamic}, a neural variational model that uses RNNs to model continuous topic evolution; and CFDTM~\cite{wu2024dynamic}, a chain-free model that replaces Markov assumptions with contrastive learning and irrelevant word exclusion.

All these models produce a temporal topic-word distribution tensor $\beta \in \mathbb{R}^{T \times K \times V}$, where $T$ is the number of timestamps, $K$ is the number of topics, and $V$ is the vocabulary size. Each entry $\beta_{t,k,v}$ denotes the probability of word $v$ under topic $k$ at time $t$. This structure forms the backbone of DTECT’s exploration and interpretation layers.

\subsection{Evaluation \& Model Selection}
To assess how well models capture temporal topic dynamics, we use three metrics from \citet{charu2024evaluating}: \textbf{Temporal Topic Coherence (TTC)}, which measures the semantic coherence of topic words over time; \textbf{Temporal Topic Smoothness (TTS)}, which quantifies how gradually topic distributions evolve across adjacent timestamps; and \textbf{Temporal Topic Quality (TTQ)}, a composite score that balances both coherence and smoothness.

Figure~\ref{fig:code_training} shows how DTECT integrates these into training and model selection, ensuring consistent comparisons. Section~\ref{sec:quantitative} presents a detailed analysis using these metrics to select the best model per corpus.

\section{Exploration \& Interpretation Layer}

DTECT provides an interactive \href{https://streamlit.io/}{Streamlit} interface, linking dynamic topic model outputs with a suite of exploration tools, including topic labeling, identification, and visualization of temporally informative words, document retrieval, summarization, and follow-up analysis. It supports various LangChain-compatible models, including those from OpenAI, Gemini, and Anthropic.

\subsection{Topic Labeling} \label{sec:labeling}
Topic models output word distributions without semantic labels, requiring manual inspection of top words for interpretation. This becomes especially challenging in dynamic settings with many topics over time. To address this, DTECT builds a \textit{temporal keyword trajectory} by extracting the top-$N$ words at each time point, e.g.,

\begin{tcolorbox}[
  colback=blue!3,
  colframe=blue!60,
  boxrule=1pt,
  rounded corners,
  left=-2mm,
  right=-2mm,
  top=0.5mm,
  bottom=0.5mm,
  width=\linewidth
]
\footnotesize
\begin{adjustbox}{width=\linewidth}
\begin{tabular}{r l}
\texttt{2019:} & \texttt{outbreak, pneumonia, china, wuhan, virus} \\
\texttt{2020:} & \texttt{lockdown, pandemic, covid, quarantine, mask} \\
\texttt{2021:} & \texttt{vaccine, immunity, doses, pfizer, rollout} \\
\end{tabular}
\end{adjustbox}
\end{tcolorbox}

This trajectory is sent to an LLM to generate a concise, descriptive label capturing the evolving theme (see Appendix~\ref{appendix:llm-labeling-prompt}). To avoid redundant API calls, results are cached using SHA-256 hashes.

\begin{table*}[ht]
\centering
\begin{adjustbox}{width=\linewidth}
\begin{tabular}{@{}lll@{}} % @{} removes extra space at the edges
\toprule
\textbf{Component} & \textbf{Subcomponent} & \textbf{Implementation} \\
\midrule

\multirow{3}{*}{\textbf{Data-to-Model Pipeline}}
& Preprocessing & OCTIS \cite{terragni2021octis}, \href{https://radimrehurek.com/gensim/models/phrases.html}{Gensim Phrases} \\
& Model Training & DTM \cite{blei2006dynamic}, DETM \cite{dieng2019dynamic}, CFDTM \cite{wu2024dynamic} \\
& Evaluation & TTC, TTS, TTQ \cite{charu2024evaluating} \\

\midrule

\multirow{5}{*}{\makecell[l]{\textbf{Exploration \&}\\\textbf{Interpretation Layer}}}
& Topic Labeling & LLM-based (GPT, Claude, Gemini) \\
& Trend Visualization & Burstiness; Specificity; Uniqueness; Time-series plots using \href{https://plotly.com/}{Plotly} \\
& \makecell[l]{Document Retrieval\\ \& Summarization} & Retrieve by Word-Timestamp; Highlight Relevant Texts; Summarize via LLMs \\
& Chat Assistant & Grounded in Retrieved Docs; Context-Aware; Factually Consistent \\

\bottomrule
\end{tabular}
\end{adjustbox}
\caption{The architecture of DTECT, outlining its primary components, their sub-components, and the corresponding implementation details.}
\label{tab:dtect_components}
\end{table*}

\subsection{Temporally Informative Word Detection and Trend Visualization} \label{sec:suggest_words}
Top-ranked topic words are often stable over time, making them less effective for capturing evolving trends. Identifying temporally salient words manually is also difficult without domain expertise. For instance, in the context of \textit{India’s financial discourse}:

\begin{tcolorbox}[
  colback=blue!3,
  colframe=blue!60,
  boxrule=1pt,
  rounded corners,
  left=-2mm,
  right=-2mm,
  top=0.5mm,
  bottom=0.5mm,
  width=\linewidth
]
\footnotesize
\begin{adjustbox}{width=\linewidth}
\begin{tabular}{r l}
\texttt{2015:} & \textit{bank}, \textit{rbi}, \textit{issue}, \texttt{cash}, \texttt{debit\_card} \\
\texttt{2016:} & \textit{rbi}, atm, \textit{issue}, \texttt{demonetisation,} \texttt{black\_money} \\
\texttt{2017:} & \textit{issue}, \textit{rbi}, \texttt{digital\_payment}, \texttt{upi}, \texttt{npci} \\
\end{tabular}
\end{adjustbox}
\end{tcolorbox}

\noindent Common terms like \textit{bank} or \textit{rbi} remain static, while emerging words like \texttt{demonetisation} and \texttt{upi} highlight major events such as the \href{https://en.wikipedia.org/wiki/2016_Indian_banknote_demonetisation}{\textbf{2016 demonetisation}} and the rise of digital payments.

To automatically identify temporally significant terms, we score each word $v$ in a topic $k$ based on its \textit{probability trajectory over time}, $\{\beta_{t,k,v}\}_{t=1}^T$. The score combines three criteria:

\paragraph{Burstiness ($S_{\text{burst}}$):} Captures temporal dynamism by comparing a word's peak probability to its mean within the topic: 
\[
    S_{\text{burst}}(v, k) = \frac{\max_{t} \beta_{t,k,v}}{\text{mean}_{t} \beta_{t,k,v} + \epsilon}
\]

\paragraph{Specificity ($S_{\text{spec}}$):}  Measures a word’s peak relevance in the topic relative to its average relevance across all topics and timestamps:
\[
    S_{\text{spec}}(v, k) = \frac{\max_{t} \beta_{t,k,v}}{\text{mean}_{t,k'} \beta_{t,k',v} + \epsilon}
\]

\paragraph{Uniqueness ($S_{\text{uniq}}$):} An IDF-style score that penalizes words active in many topics: 
\[
    S_{\text{uniq}}(v) \propto \log \left( \frac{\text{Total Topics}}{|\{k' : v \in \text{Top-}N\text{ words of } k'\}|} \right)
\]

\noindent The final interestingness score is the product of these three components:
\[
    S_{\text{final}}(v, k) = S_{\text{burst}}(v, k) \times S_{\text{spec}}(v, k) \times S_{\text{uniq}}(v)
\]
\noindent where $\epsilon$ is a small positive constant added for numerical stability.

Ranking words by $S_{\text{final}}$ surfaces key terms that signal temporal shifts. Their evolution can be visualized by plotting their probability trajectories, $\text{trend}(v) = [\beta_{1,k,v}, \dots, \beta_{T,k,v}]$.

\subsection{Document Retrieval and Summarization} \label{sec:summery}
DTECT enables users to explore any word-timestamp pair by retrieving documents from the corresponding time that contain the selected word. To ensure both relevance and diversity, we first retrieve candidate documents using \textsc{Faiss}, then apply Maximal Marginal Relevance (MMR) to refine the selection. The final set is synthesized into a thematic, bullet-point summary by an LLM. The full prompt is provided in Appendix~\ref{appendix:llm-summary-prompt}.

\subsection{Conversational Exploration} \label{sec:chat}
To enable deeper engagement with the summarized content, DTECT provides a chat interface for follow-up questions, implemented via LangChain’s \texttt{ConversationChain} to maintain dialogue history. Each response is strictly grounded in the retrieved documents by prepending a system prompt that instructs the LLM to rely only on the provided content and indicate when information is missing. This ensures factual consistency and makes responses directly interpretable. The system prompt is included in Appendix~\ref{appendix:llm-chat-prompt}.

To summarize the full system, Table~\ref{tab:dtect_components} outlines the architecture of DTECT, highlighting its two main components—\textit{Data-to-Model Pipeline} and \textit{Exploration \& Interpretation Layer}—along with the models, tools, and techniques integrated at each stage.

\begin{figure*}[!ht]
    \centering
    \fbox{\includegraphics[width=.65\linewidth]{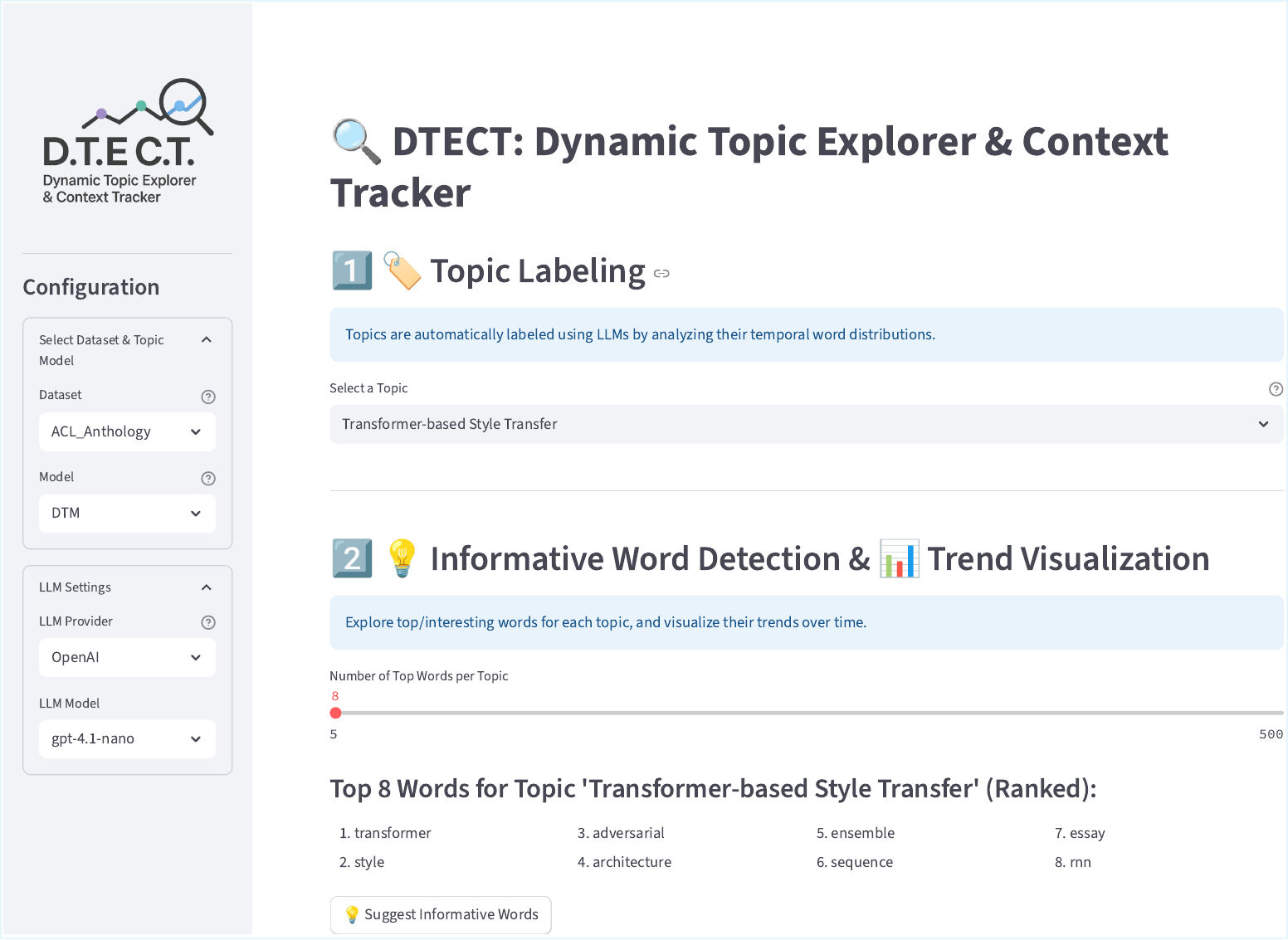}}
    \caption{Overview of the DTECT interface. After selecting the dataset, model, and LLM settings, users can begin exploring topics by choosing a topic from the list of LLM-generated topic-labels.}
    \label{fig:overview}
\end{figure*}

\section{User Interface}
DTECT offers an interactive web-based interface, available both locally and at \url{https://huggingface.co/spaces/AdhyaSuman/DTECT}, with source code under the MIT license and a demo video at \url{https://youtu.be/B8nNfxFoJAU}. Figure~\ref{fig:overview} shows an overview.

Users begin by preprocessing their dataset (Figure~\ref{fig:code-preprocessing}), followed by model training and evaluation (Figure~\ref{fig:code_training}). Once completed, they can interactively explore results by selecting the dataset, model, and LLM settings. DTECT then generates and caches the inverted index and topic labels for efficient reuse (\textsection\ref{sec:labeling}).

Topic exploration begins with selecting a topic and identifying temporally salient words (\textsection\ref{sec:suggest_words}), visualized via interactive \href{https://plotly.com/}{Plotly} charts. Clicking a word-timestamp pair retrieves matching documents, which can be summarized (\textsection\ref{sec:summery}) or further examined through a document-grounded chat interface (\textsection\ref{sec:chat}). All components operate in real-time, enabling seamless, end-to-end exploration.

\section{Evaluation}
We validate DTECT through comparative analysis with existing tools, quantitative evaluation of dynamic topic models, user feedback, and a real-world case study.

\subsection{Comparison to Existing Toolkit}
Toolkits like \textsc{Gensim} \cite{rehurek2010gensim} and \textsc{OCTIS} \cite{terragni2021octis} focus on static topic modeling. \textsc{TopMost} \cite{wu2024topmost} supports dynamic models but lacks interpretability and exploration features.

\begin{table}[!ht]
    \centering
    \begin{adjustbox}{width=\linewidth}
      \begin{tabular}{|l|c|c|c|}
        \toprule
        \textbf{Features} & {DTECT} & \textsc{TopMost} & \textsc{Gensim}\\
        \midrule
        \makecell[l]{Availability of models:\\ 1. DTM} & \makecell[c]{\\ \textcolor{ao}{\cmark}} & \makecell[c]{\\ \textcolor{ao}{\cmark}} &  \makecell[c]{\\ \textcolor{ao}{\cmark}}\\
        2. DETM & \textcolor{ao}{\cmark} & \textcolor{ao}{\cmark} & \textcolor{red}{\xmark}\\
        3. CFDTM & \textcolor{ao}{\cmark} & \textcolor{ao}{\cmark} & \textcolor{red}{\xmark}\\ \midrule
        % 4. DBERTopic & \textcolor{ao}{\cmark} & \textcolor{red}{\xmark} & \textcolor{red}{\xmark}\\ \midrule
        
        Temporal evaluation & \textcolor{ao}{\cmark} & \textcolor{ao}{\cmark} & \textcolor{red}{\xmark} \\
        Topic labeling & \textcolor{ao}{\cmark} & \textcolor{red}{\xmark} & \textcolor{red}{\xmark} \\
        Informative terms detection & \textcolor{ao}{\cmark} & \textcolor{red}{\xmark} & \textcolor{red}{\xmark} \\
        Interactive UI & \textcolor{ao}{\cmark} & \textcolor{red}{\xmark} & \textcolor{red}{\xmark}\\
        Summarization & \textcolor{ao}{\cmark} & \textcolor{red}{\xmark} & \textcolor{red}{\xmark}\\
        Chat assistant & \textcolor{ao}{\cmark} & \textcolor{red}{\xmark} & \textcolor{red}{\xmark}\\
        \bottomrule
      \end{tabular}
    \end{adjustbox}
    \caption{Feature comparison of DTECT with \textsc{TopMost} and \textsc{Gensim}}
    \label{tab:dtect_topmost}
\end{table}

As shown in Table~\ref{tab:dtect_topmost}, DTECT supports multiple models (DTM, DETM, CFDTM) and integrates key features: dynamic evaluation metrics (TTC, TTS, TTQ), LLM-based labeling, keyword tracking, interactive UI, summarization, and a chat assistant, making it a comprehensive platform for dynamic topic modeling.

\subsection{Quantitative Evaluation of Dynamic Topic Models} \label{sec:quantitative}
We evaluated the performance of DTECT's models (DTM, DETM, and CFDTM) on three datasets: ACL Anthology, UN Debates, and TCPD-IPD Finance. Table~\ref{tab:topic_quality} reports topic quality scores (TTC, TTS, TTQ) and runtimes. The results show that while the classic DTM yields the highest-quality topics, its computational cost is substantial. DETM, in contrast, offers a compelling balance of efficiency and competitive performance, making it a practical choice for most applications.

\begin{table}[ht]
    \centering
    \begin{adjustbox}{width=\linewidth}
    \begin{tabular}{l c ccc}
        \toprule
        \textbf{Model} & \textbf{Time} (Sec.) $\downarrow$ & \textbf{TTC} $\uparrow$ & \textbf{TTS} $\uparrow$ & \textbf{TTQ} $\uparrow$ \\ \midrule
        \multicolumn{5}{c}{\textbf{ACL Anthology}} \\ \midrule
        CFDTM     & 18,383 & 0.0751 & 0.9127 & 0.0785 \\
        DETM      & \textbf{13,638} & 0.1006 & 0.8448 & 0.0838 \\
        DTM       & 73,992 & \textbf{0.1050} & \textbf{0.9430} & \textbf{0.0981} \\
        \midrule
        \multicolumn{5}{c}{\textbf{UN Debates}} \\\midrule
        CFDTM     & 7,450 & -0.0016 & 0.5832 & 0.0032 \\
        DETM      & \textbf{4,792} & 0.0980  & 0.8721 & 0.0858 \\
        DTM       & 2,90,400 & \textbf{0.1321} & \textbf{0.9386} & \textbf{0.1239} \\
        \midrule
        \multicolumn{5}{c}{\textbf{TCPD-IPD Finance}} \\ \midrule
        CFDTM     & 6,625 & 0.1533 & 0.7843 & 0.1300 \\
        DETM      & \textbf{6,017} & 0.1449 & 0.7883 & 0.1173 \\
        DTM       & 1,17,468 & \textbf{0.1818} & \textbf{0.9505} & \textbf{0.1731} \\
        \bottomrule
    \end{tabular}
    \end{adjustbox}
    \caption{Topic quality and runtime comparison across datasets.}
    \label{tab:topic_quality}
\end{table}

\subsection{User Feedback Evaluation}
We collected feedback from 25 users who watched a \href{https://youtu.be/B8nNfxFoJAU}{demo video} of DTECT and rated seven aspects of the system on a 5-point Likert scale. As shown in Table~\ref{tab:human_eval}, all components received high ratings (mean > 4.5) with low variance, reflecting consistent positive user perception.

\begin{table}[!ht]
\centering
\begin{adjustbox}{width=.88\linewidth}
    \begin{tabular}{lcc}
        \toprule
        \textbf{Evaluation Aspect} & \textbf{Mean} & \textbf{Std.} \\
        \midrule
        Ease of Navigation          & 4.64 & 0.57 \\
        Tool Responsiveness         & 4.52 & 0.59 \\
        Topic Labeling              & 4.64 & 0.57 \\
        Informative Word Suggestion & 4.68 & 0.56 \\
        Document Retrieval          & 4.68 & 0.56 \\
        Generated Summary           & 4.60 & 0.71 \\
        Chat Assistant              & 4.48 & 0.77 \\
        \bottomrule
    \end{tabular}
\end{adjustbox}
\caption{Statistics of user ratings (1–5) for DTECT components based on feedback from 25 users.}
\label{tab:human_eval}
\end{table}

\begin{figure}[!ht]
    \centering
    \fbox{\includegraphics[width=\linewidth]{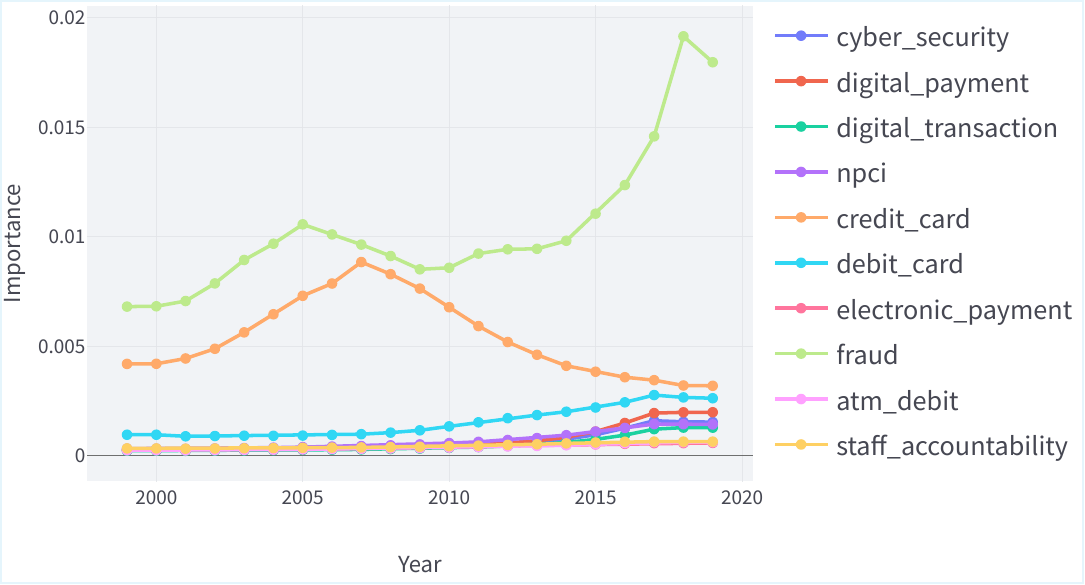}}
    \caption{Temporal Trends of Informative Words for the Topic \textit{Banking Regulations} within the TCPD-IPD Dataset (Finance Ministry).}
    \label{fig:trend_bank}
\end{figure}

\subsection{Case Study: Parliamentary Q\&A}
We demonstrate DTECT's capabilities on the TCPD-IPD dataset, analyzing parliamentary questions to India's Ministry of Finance using the DTM model, which showed the best coherence.

\paragraph{Informative Term Detection for ``Banking Regulations":}
We explored the topic of \textit{Banking Regulations} using DTECT's {Informative Word Detection} module (\textsection\ref{sec:suggest_words}). DTECT automatically identified the top 10 informative terms from the top 500 words: \textit{``cyber\_security"}, \textit{``digital\_payment"}, \textit{``digital\_transaction"}, \textit{``npci"}, \textit{``credit\_card"}, \textit{``debit\_card"}, \textit{``electronic\_payment"}, \textit{``fraud"}, \textit{``atm\_debit"}, and \textit{``staff\_accountability"}. This automated output offers immediate interpretability, contrasting with the manual curation of over 350 words by \citet{adhya2022indian} to derive similar terms like \textit{``npci"}, \textit{``cheque"}, \textit{``debit\_card"}, \textit{``credit\_card"}, \textit{``digital\_transaction"}, \textit{``digital\_payment"}, \textit{``cyber\_security"}, and \textit{``atm"}. DTECT thus streamlines keyword identification through automated ranking.

\begin{figure}[!ht]
    \centering
    \fbox{\includegraphics[width=\linewidth]{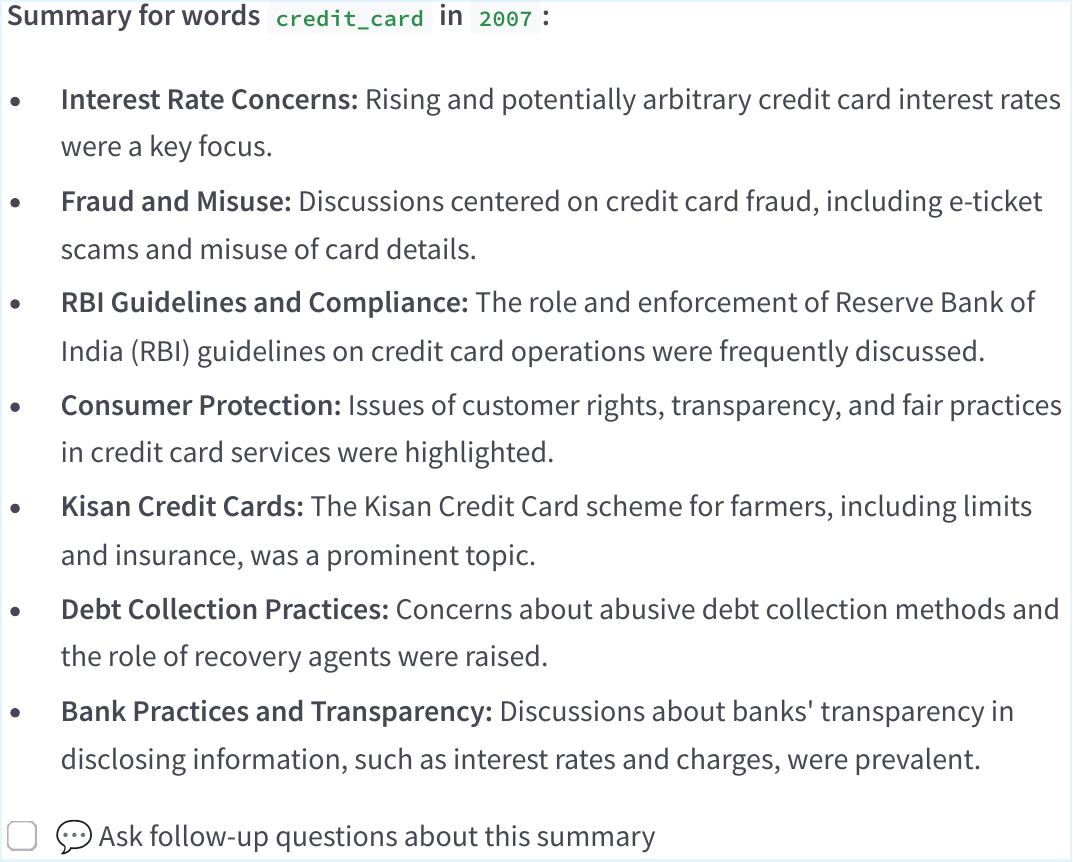}}
    \caption{Summaries generated by DTECT for the ``Credit Card" topic, highlighting key events and discussions around the 2007 peak.}
    \label{fig:summary_creditcard}
\end{figure}

\paragraph{Temporal Analysis: ``Credit Card" Spike:}
Figure~\ref{fig:trend_bank} shows temporal trends of identified keywords. Consistent with \citet{adhya2022indian}, we focus on the spike in interest around \textit{``credit\_card''} in 2007. Using DTECT's trend visualization and summarization modules, we examined the 2007 spike in \textit{``credit\_card''} interest. DTECT's trend visualization and summarization modules confirmed three main factors: (1) increased credit card frauds; (2) rising bank interest rates; and (3) expansion of the Kisan Credit Card (KCC) scheme. Figure~\ref{fig:summary_creditcard} illustrates DTECT's ability to capture these events via document-grounded summaries.

This case study highlights DTECT's support for end-to-end, interpretable dynamic topic analysis. Additional case studies are in Appendix~\ref{app:case_study}.

\section{Relevance of DTECT in the Era of LLMs} \label{sec:relevance}
While dynamic topic models are well-suited for capturing topic evolution in large corpora, LLMs face limitations due to their fixed context windows, making them less effective for long-range temporal analysis. \textbf{DTECT} offers a hybrid solution: it uses a DTM backbone to model temporal dynamics and integrates LLMs to address key usability and interpretability gaps. Specifically, DTECT employs LLMs for (1) generating interpretable topic labels, (2) summarizing document-grounded explanations for shifts in topic distributions, and (3) enabling interactive, conversational exploration of trends. This synergy reduces the need for manual inspection and streamlines dynamic topic analysis at scale.

\section{Conclusion}\label{sec:conclusion}
Navigating evolving themes in large text collections is challenging. We introduce \textbf{DTECT}, an interactive system that combines dynamic topic models for scalable analytics with LLMs for nuanced understanding. \textbf{DTECT} offers a unified interface, allowing users to effortlessly explore automatically labeled topic trends and delve into the underlying documents. Features like one-click temporal summarization and a conversational assistant streamline analysis, transforming data into an intuitive, exploratory experience that reveals the temporal narratives within any large text corpus.

\section*{Limitations}
DTECT is built as a modular and extensible framework that supports the full dynamic topic modeling pipeline, from preprocessing to interactive analysis. While the system is robust and adaptable, there remain opportunities for refinement.

The integration of LLMs for topic labeling, summarization, and conversational exploration greatly enhances interpretability. However, like most LLM-based components, these features introduce challenges related to transparency, reproducibility, and sensitivity to domain shifts. To mitigate these concerns, DTECT incorporates several mechanisms to promote user trust and validation. For topic labeling, the system displays top words alongside generated labels, allowing users to assess their semantic coherence. For summarization, the MMR-selected supporting documents are made visible, and all prompt templates used in generation are detailed in Appendix~\ref{appendix:llm-prompts}, ensuring transparency and contextual grounding.

Currently, DTECT supports a core set of dynamic topic models. Although not exhaustive, its model-agnostic design allows easy integration of additional methods. Expanding model coverage remains a priority for future development.

% \bibliography{custom}
% \input{output.bbl}

\appendix

\section{LLM Prompts}\label{appendix:llm-prompts}

This section provides the templates for the prompts used in our system for topic labeling, summarization, and conversational follow-up. All prompts are executed with the language model temperature set to \textbf{0} to ensure consistency and reproducibility across outputs.

\subsection{Topic Labeling Prompt Template}\label{appendix:llm-labeling-prompt}
This prompt guides the LLM to assign concise and semantically meaningful labels to topics based on their top words over time, improving the interpretability of dynamic topic models.

\begin{tcolorbox}[
    title=\centering \textbf{Prompt for Topic Labeling},
    colback=blue!3,
    colframe=blue!60,
    coltitle=white,
    fonttitle=\bfseries,
    rounded corners,
    boxrule=0.6pt,
    width=\linewidth
]
You are an expert in topic modeling and temporal data analysis. Given the top words for a topic across multiple time points, your task is to return a short, specific, descriptive topic label. Avoid vague, generic, or overly broad labels. Focus on consistent themes in the top words over time. Use concise noun phrases, 2--5 words max. Do \textbf{not} include any explanation, justification, or extra output.

\medskip
\noindent\textbf{Top words over time:} \{\texttt{trajectory}\}

\medskip
\noindent\textbf{Return ONLY the label (no quotes, no extra text):}
\end{tcolorbox}

\subsection{Summarization Prompt Template} \label{appendix:llm-summary-prompt}
The summarization module uses the following prompt to generate concise thematic summaries of selected documents:

\begin{tcolorbox}[
    title=\centering \textbf{Prompt for Thematic Summarization},
    colback=blue!3,
    colframe=blue!60,
    coltitle=white,
    fonttitle=\bfseries,
    rounded corners,
    boxrule=0.6pt,
    width=\linewidth
]
Given the following documents from \{\texttt{timestamp}\} that mention the words: \{\texttt{word\_list}\}, identify the key themes or discussion points from that time. Be concise. Each bullet should capture a distinct theme in 1-2 short sentences. Avoid any elaboration, examples, or justification.

\medskip
Return no more than 5-7 bullets.

\medskip
\{\texttt{context\_texts}\}

\medskip
\textbf{Summary:}
\end{tcolorbox}

\subsection{Conversational Follow-up System Instruction} \label{appendix:llm-chat-prompt}
To ensure document-grounded answers during Q\&A, the following system instruction is used:

\begin{tcolorbox}[
    title=\centering \textbf{Prompt for Chat},
    colback=blue!3,
    colframe=blue!60,
    coltitle=white,
    fonttitle=\bfseries,
    rounded corners,
    boxrule=0.6pt,
    width=\linewidth
]
You are an assistant answering questions strictly based on the provided sample documents below.

If the answer is not clearly supported by the text, respond with: \textit{``The information is not available in the documents provided."}

\medskip
\noindent\textbf{Documents:} \{\texttt{context\_texts}\}

\medskip
\noindent\textbf{User Question:} \{\texttt{user\_question}\}
\end{tcolorbox}

\section{Supporting Resources}

We list below the datasets, codebases, and evaluation resources referenced or integrated into DTECT:

\paragraph{Datasets.}
\begin{itemize}
    \item \textbf{ACL Anthology:} \url{https://aclanthology.org/}
    \item \textbf{UN General Debates:} \url{https://www.kaggle.com/datasets/unitednations/un-general-debates}
    \item \textbf{TCPD-IPD Finance:} \url{https://tcpd.ashoka.edu.in/question-hour/}
\end{itemize}

\paragraph{Dynamic Topic Modeling Codebases.}
\begin{itemize}
    \item \textbf{TopMost Toolkit:} \url{https://github.com/bobxwu/TopMost}
    \item \textbf{DTM Implementation:} \url{https://github.com/bobxwu/TopMost/blob/main/topmost/trainers/dynamic/DTM_trainer.py}
    \item \textbf{DETM Implementation:} \url{https://github.com/bobxwu/TopMost/blob/main/topmost/models/dynamic/DETM.py}
    \item \textbf{CFDTM Implementation:} \url{https://github.com/bobxwu/TopMost/tree/main/topmost/models/dynamic/CFDTM}
\end{itemize}

\paragraph{Preprocessing Toolkit.}
\begin{itemize}
    \item \textbf{OCTIS:} \url{https://github.com/MIND-Lab/OCTIS}
\end{itemize}

\paragraph{Evaluation Metrics.}
\begin{itemize}
    \item \textbf{Evaluating Dynamic Topic Models:} \url{https://github.com/CharuJames/Evaluating-Dynamic-Topic-Models}
\end{itemize}

\section{User Feedback}

To evaluate the usability and effectiveness of \textbf{DTECT}, we collected structured user feedback via a Google Form following a guided \href{https://youtu.be/B8nNfxFoJAU}{demo video} and hands-on user interaction. The form included Likert-scale questions covering the following aspects: (1) \textbf{Ease of Navigation} – assessing the intuitiveness of the user interface and overall tool flow; (2) \textbf{Tool Responsiveness} – measuring the system’s performance and interaction speed; (3) \textbf{Topic Labeling} – focusing on the clarity and relevance of the automatically generated topic labels; (4) \textbf{Informative Words} – evaluating the usefulness of suggested keywords in capturing meaningful temporal patterns; (5) \textbf{Document Retrieval} – examining the accuracy and responsiveness of the system for keyword-year based queries; (6) \textbf{Summary Quality} – reflecting how informative and coherent the generated summaries were; and (7) \textbf{Chat Assistant} – rating the relevance and accuracy of responses based on document summaries.

\begin{figure}[!ht]
    \centering
    \fbox{\includegraphics[width=\linewidth]{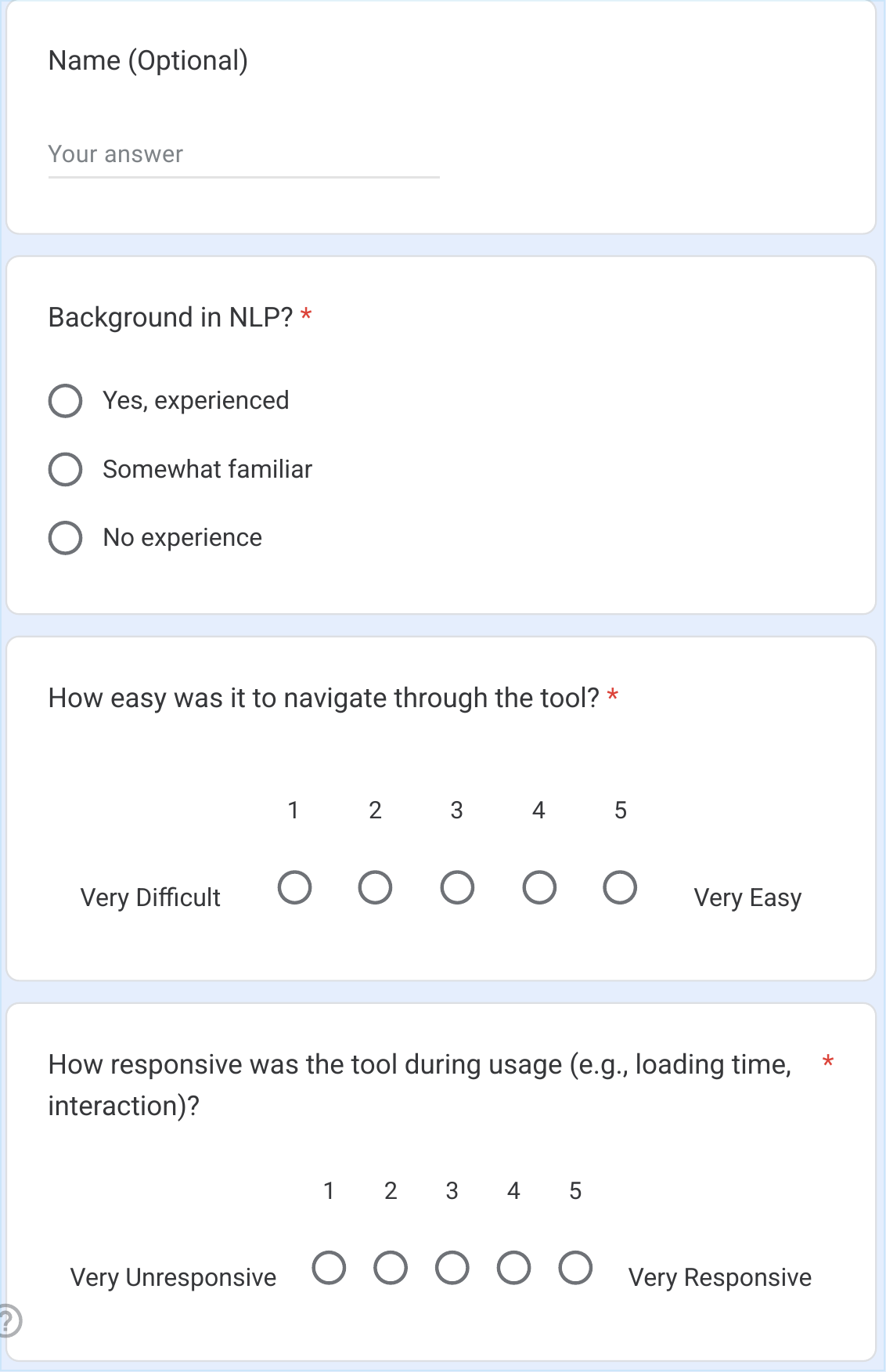}}
    \caption{User demographics and general usability ratings collected via feedback form.}
    \label{fig:feedback_1}
\end{figure}

\begin{figure}[!ht]
    \centering
    \fbox{\includegraphics[width=\linewidth]{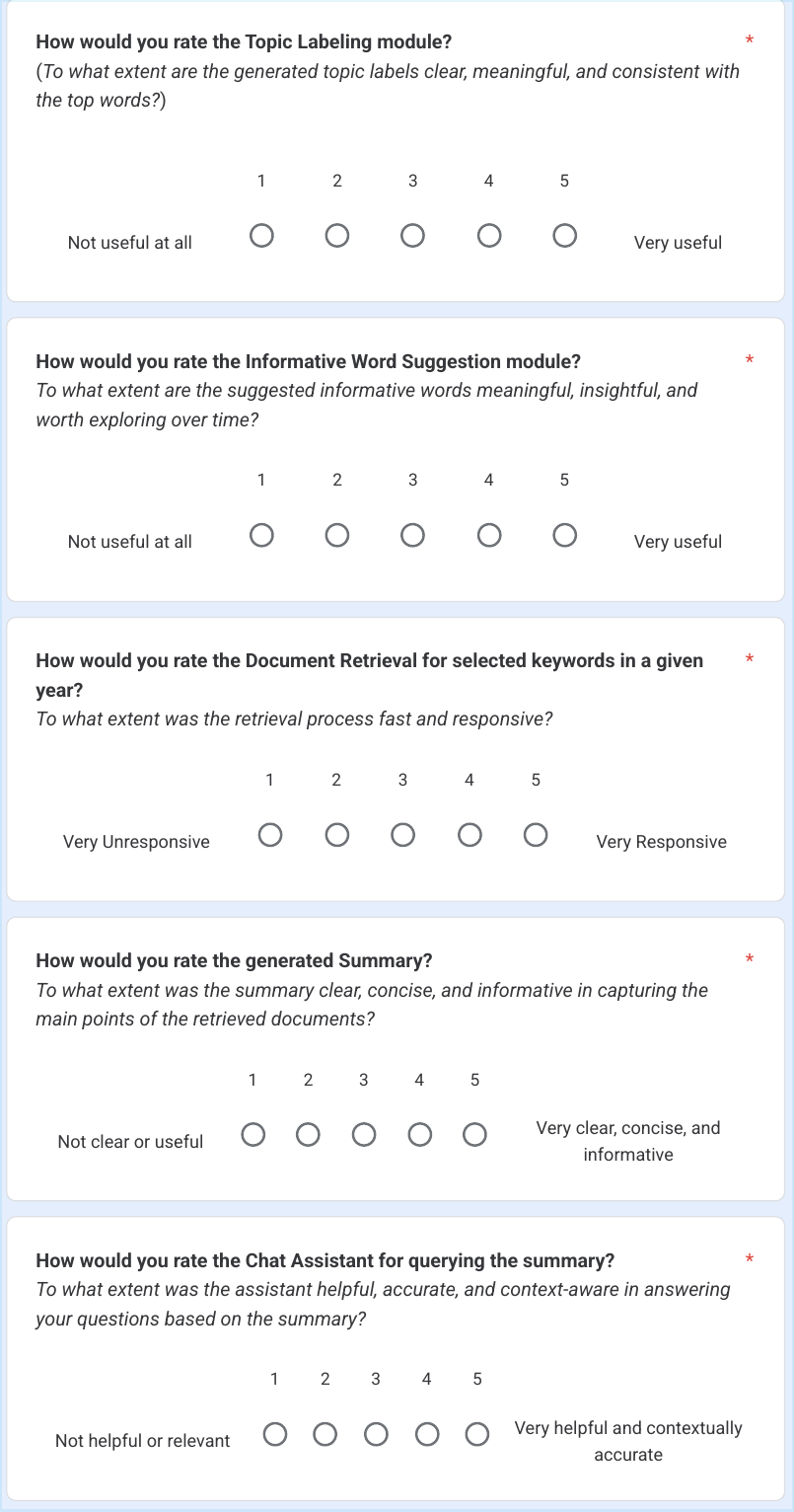}}
    \caption{Component-wise user ratings for DTECT modules including Topic Labeling, Informative Words, Retrieval, Summarization, and Chat Assistant.}
    \label{fig:feedback_2}
\end{figure}

Figures~\ref{fig:feedback_1} and~\ref{fig:feedback_2} illustrate the interface and sample questions users encountered in the feedback form.

\section{Example Case Studies} \label{app:case_study}
We present here two illustrative case studies demonstrating how DTECT enables in-depth, interpretable analysis of topic evolution across diverse domains: scientific research and international political discourse.

\subsection{Case Study: ACL Anthology (2010--2025)}
To demonstrate DTECT’s use on research corpora, we analyze the ACL Anthology (2010--2025), focusing on the topic of \textit{Neural Machine Translation} using the DTM model.

\begin{figure}[!ht]
    \centering
    \fbox{\includegraphics[width=\linewidth]{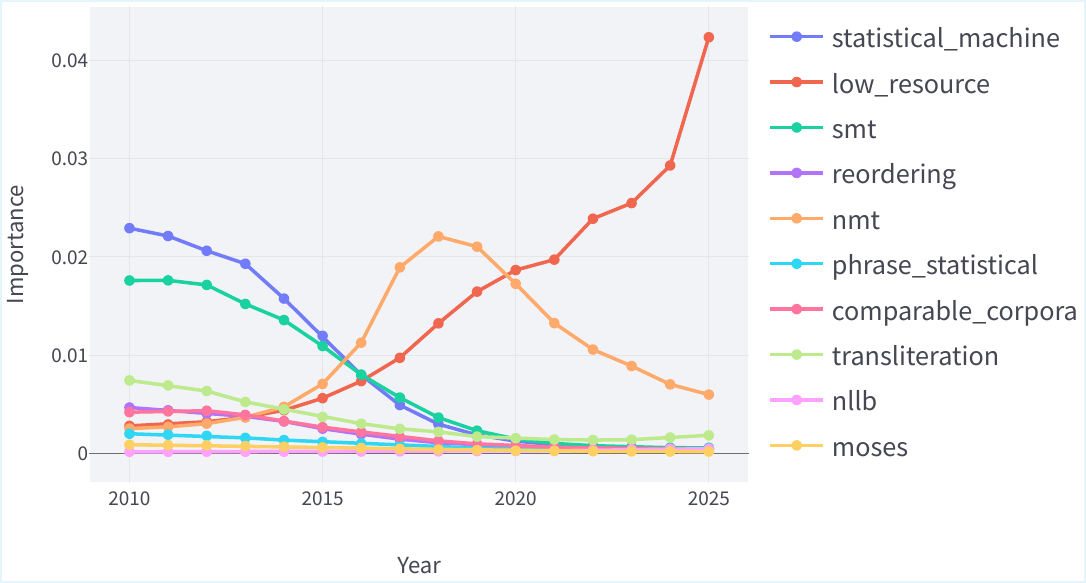}}
    \caption{Temporal trends of informative words for the topic \textit{Neural Machine Translation} in the ACL Anthology dataset.}
    \label{fig:trend_nmt}
\end{figure}

\begin{figure}[!ht]
    \centering
    \fbox{\includegraphics[width=\linewidth]{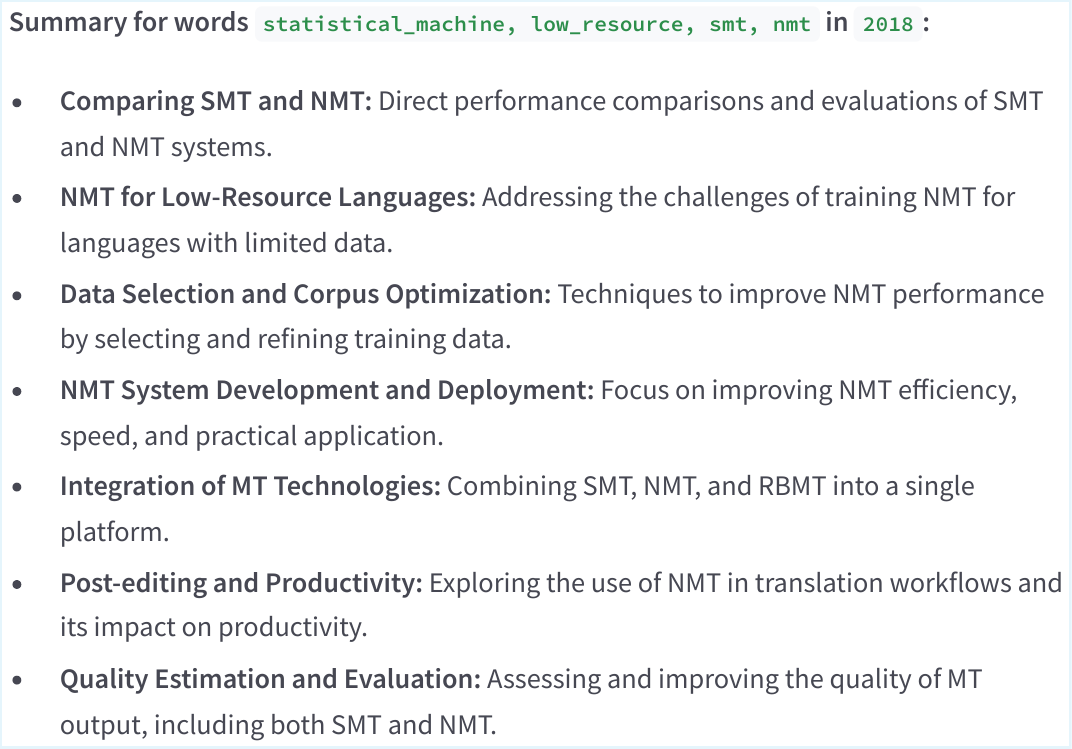}}
    \caption{Summaries generated by DTECT for 2018, highlighting SMT vs. NMT comparisons, low-resource MT, and system deployment.}
    \label{fig:summary_nmt}
\end{figure}

\paragraph{Informative Term Detection for ``Neural Machine Translation":}
DTECT’s \textit{Informative Word Detection} module (\textsection\ref{sec:suggest_words}) identifies key terms such as: \textit{``statistical\_machine"}, \textit{``low\_resource"}, \textit{``smt"}, \textit{``reordering"}, \textit{``nmt"}, \textit{``phrase\_statistical"}, \textit{``comparable\_corpora"}, \textit{``transliteration"}, \textit{``nllb"}, and \textit{``moses"}. These terms capture the transition from statistical and rule-based approaches to neural paradigms. Their temporal trends are explored below.

\paragraph{Interactive Trend Analysis and Summary Insights}
DTECT’s trend visualization shows that interest in \textit{``nmt"} peaks in 2018, reflecting the community’s strong shift toward neural approaches. In contrast, \textit{``smt"} and \textit{``statistical\_machine"} decline steadily after 2010, while \textit{``low\_resource"} gradually gains prominence, signaling a growing interest in multilingual and under-resourced language scenarios.

We focus on the year 2018 to generate document-grounded summaries using the terms \textit{``statistical\_machine"}, \textit{``low\_resource"}, \textit{``smt"}, and \textit{``nmt"} (Figure~\ref{fig:summary_nmt}). The summarization module reveals core themes such as SMT vs. NMT comparisons, challenges in low-resource translation, data selection strategies, deployment of NMT systems, integration of hybrid approaches, post-editing workflows, and quality estimation.

\begin{figure}[!ht]
    \centering
    \fbox{\includegraphics[width=\linewidth]{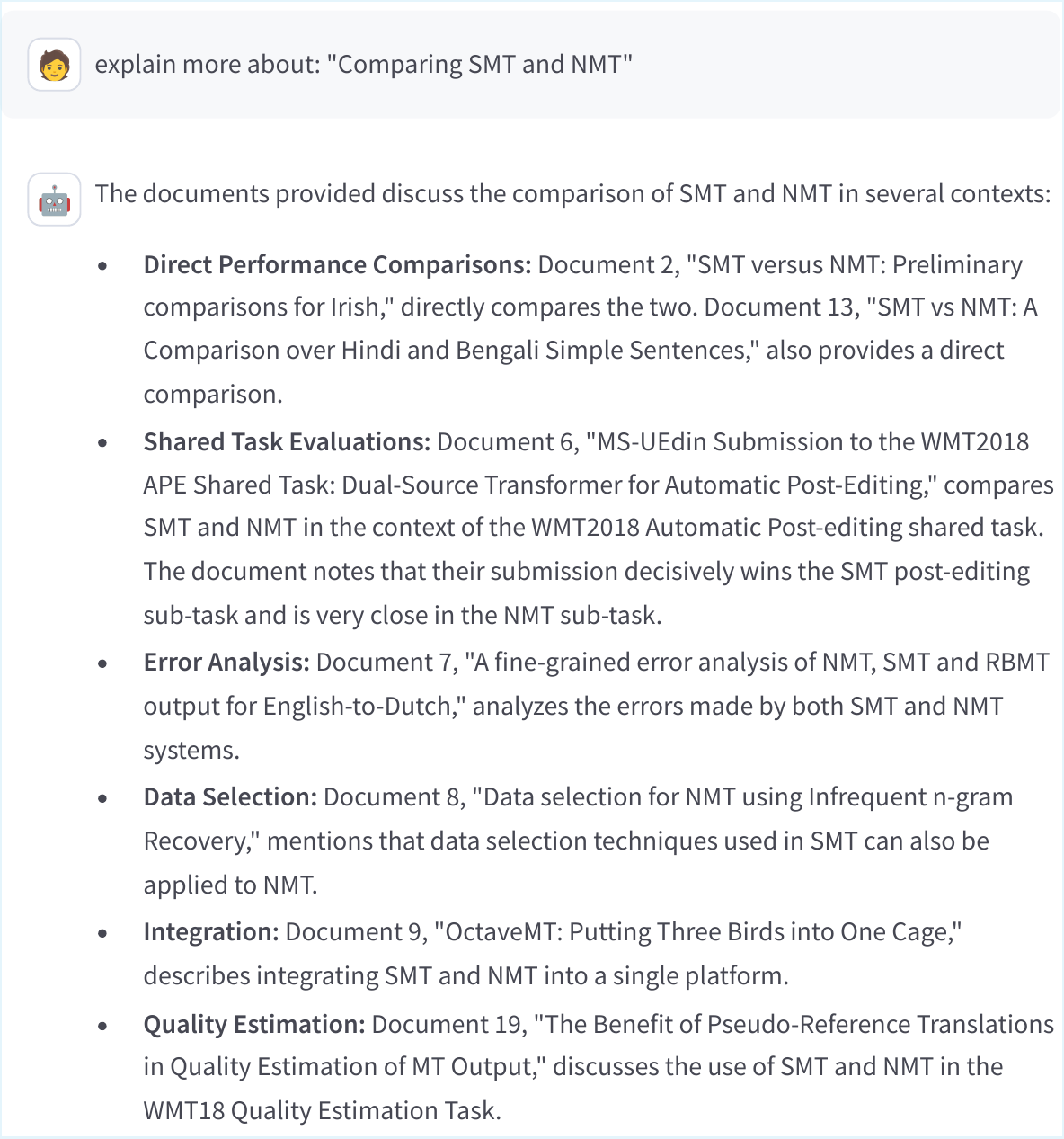}}
    \caption{DTECT’s chat assistant responds to a follow-up query on SMT vs. NMT, retrieving comparative studies across tasks and languages.}
    \label{fig:followup_nmt}
\end{figure}

\paragraph{Follow-up Exploration: SMT vs. NMT Comparisons:}
To explore this comparison further, we used DTECT’s chat interface to query \textit{SMT vs. NMT}. The system retrieved a diverse set of documents covering performance evaluation, shared tasks, and system integration.

Document 2 (“SMT versus NMT: Preliminary comparisons for Irish”) and Document 13 (“SMT vs NMT: A Comparison over Hindi and Bengali Simple Sentences”) present direct comparisons. Document 6 discusses performance in the WMT18 post-editing shared task, showing strong results for both SMT and NMT. Document 7 provides error analysis for English--Dutch. Document 8 explores how SMT-style data selection aids NMT training, while Document 9 discusses integrating SMT, NMT, and RBMT. Document 19 focuses on quality estimation across systems.

%%%%%%%%%%%%%%%%%%%%%%%%%%%%%%%%%%%%%%%%%%%%%%%%%%%%%%%%%%%%%%%%%%%
\subsection{Case Study: UN General Debates (1970--2015)}

To demonstrate DTECT’s utility for political discourse analysis, we analyze the UN General Debates dataset spanning 1970 to 2015. We focus on the topic of \textit{Global Politics}, using the DTM model selected based on coherence scores.

\begin{figure}[!ht]
    \centering
    \fbox{\includegraphics[width=\linewidth]{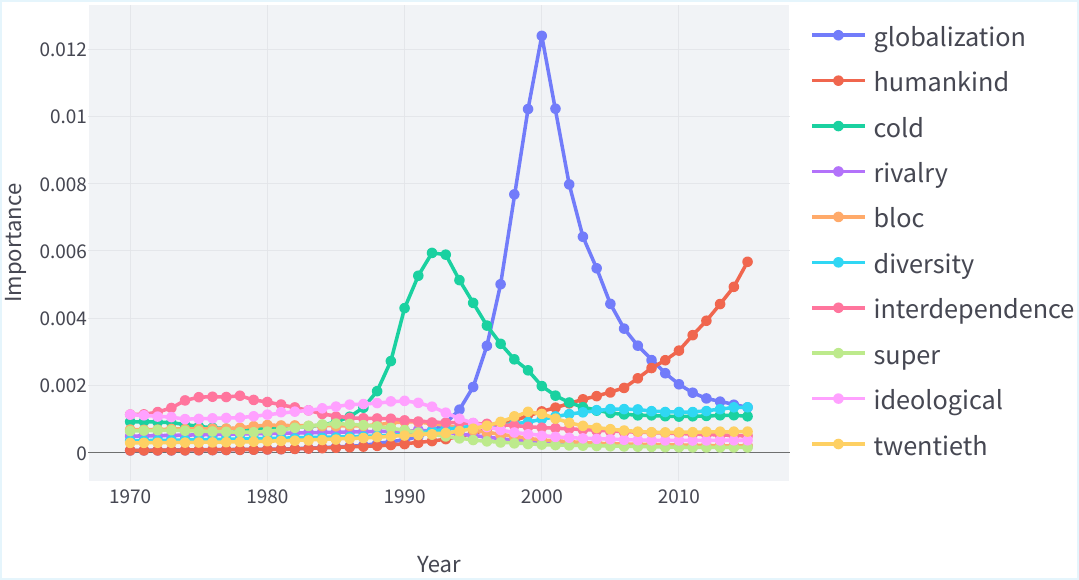}}
    \caption{Temporal trends of informative words for the topic \textit{Global Politics} in the UN General Debates dataset.}
    \label{fig:trend_global}
\end{figure}

\paragraph{Informative Term Detection for ``Global Politics":}
DTECT’s \textit{Informative Word Detection} module (\textsection\ref{sec:suggest_words}) identifies key terms for this topic: \textit{``globalization"}, \textit{``humankind"}, \textit{``cold"}, \textit{``rivalry"}, \textit{``bloc"}, \textit{``diversity"}, \textit{``interdependence"}, and others. These reflect both late 20th-century ideological tensions and growing emphasis on international cooperation. Their usage patterns over time are examined in Figure~\ref{fig:trend_global}.

\begin{figure}[!ht]
    \centering
    \fbox{\includegraphics[width=\linewidth]{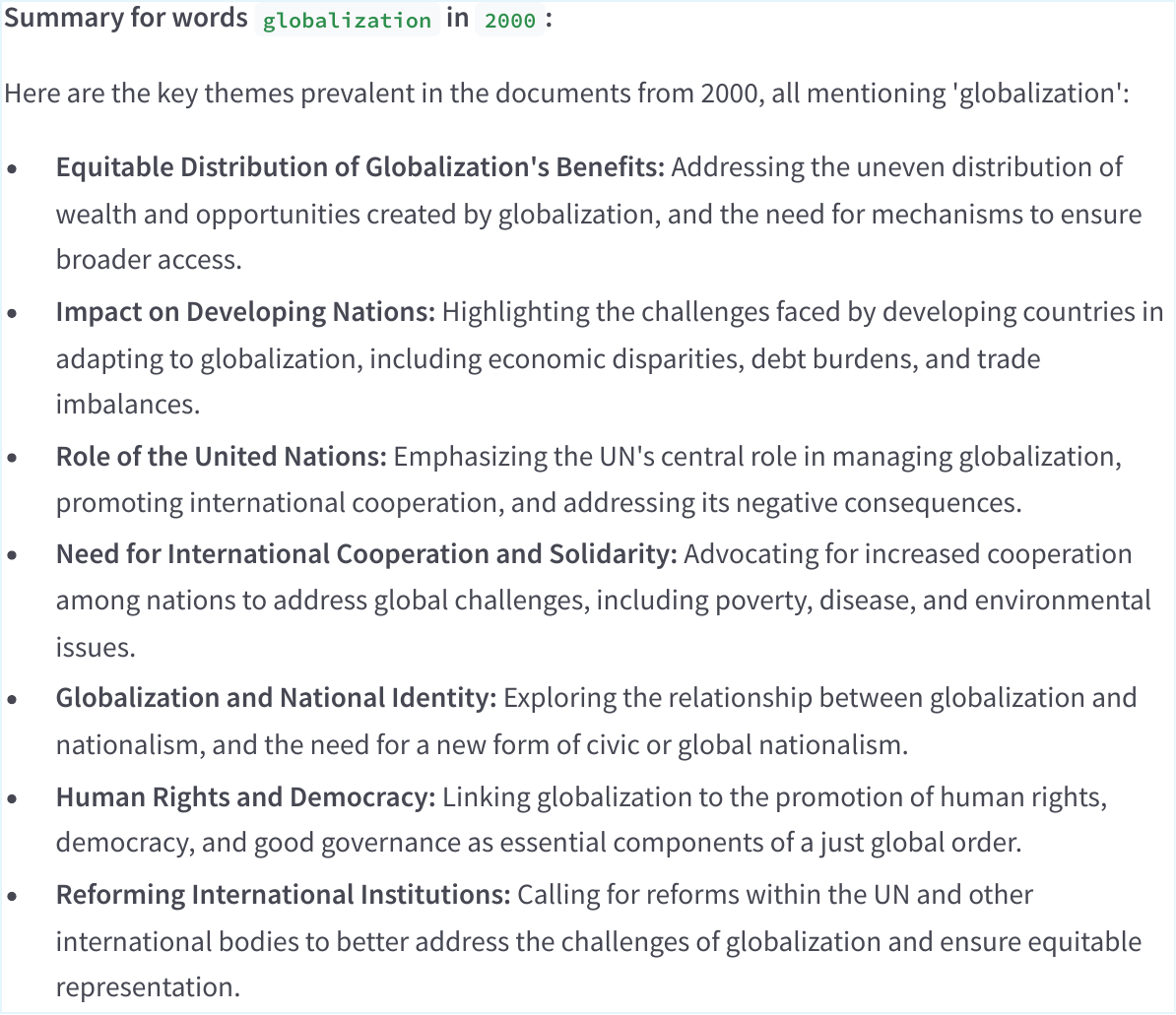}}
    \caption{Summaries generated by DTECT for the year 2000, centered on debates involving globalization.}
    \label{fig:summary_globalization}
\end{figure}

\paragraph{Interactive Trend Analysis and Summary Insights}
The term \textit{``globalization"} peaks in 2000, reflecting widespread discourse on its impact and the role of the United Nations. In that year, DTECT retrieved 143 matching documents. Another prominent term, \textit{``cold"}, peaks in 1992 with 130 documents, indicating retrospective discussion on the Cold War and its aftermath.

For 2000, using the term \textit{``globalization"}, DTECT’s summarization module reveals key concerns such as equitable distribution of globalization's benefits, its adverse impact on developing countries, the UN’s role in mediating these effects, and the call for institutional reforms (Figure~\ref{fig:summary_globalization}). Other major themes include global solidarity, nationalism, human rights, and democratic governance.

\begin{figure}[!ht]
    \centering
    \fbox{\includegraphics[width=\linewidth]{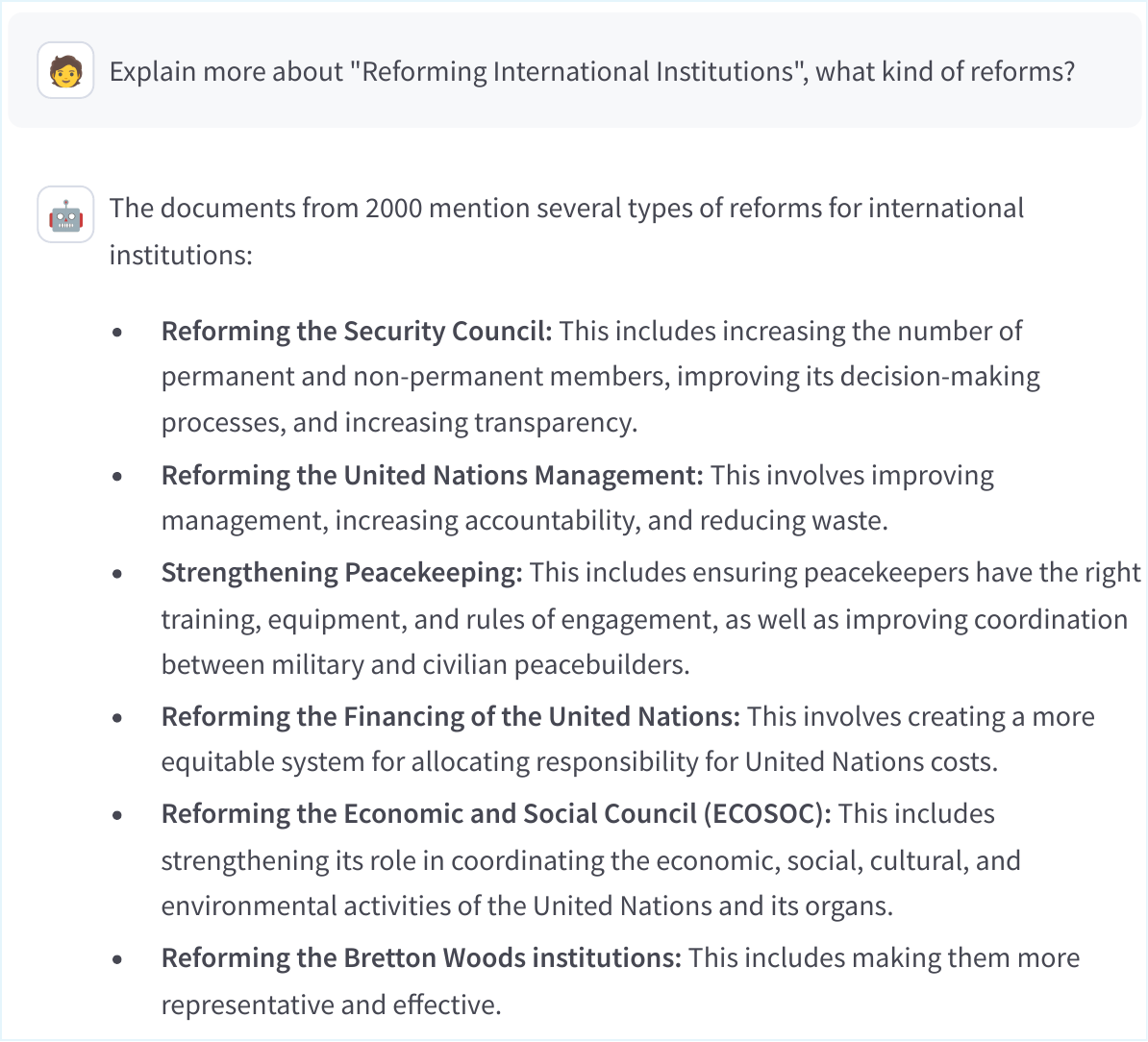}}
    \caption{DTECT’s chat assistant responds to a follow-up query on reforming international institutions.}
    \label{fig:followup_globalization}
\end{figure}

\paragraph{Follow-up Exploration: Reforming International Institutions}
A follow-up query on “Reforming International Institutions” retrieves detailed proposals discussed in the year 2000 as shown in Figure~\ref{fig:followup_globalization}. These include expanding the UN Security Council, improving peacekeeping capacity, enhancing financial accountability, and reforming ECOSOC and Bretton Woods institutions to increase global representativeness and efficiency.

\begin{figure}[!ht]
    \centering
    \fbox{\includegraphics[width=\linewidth]{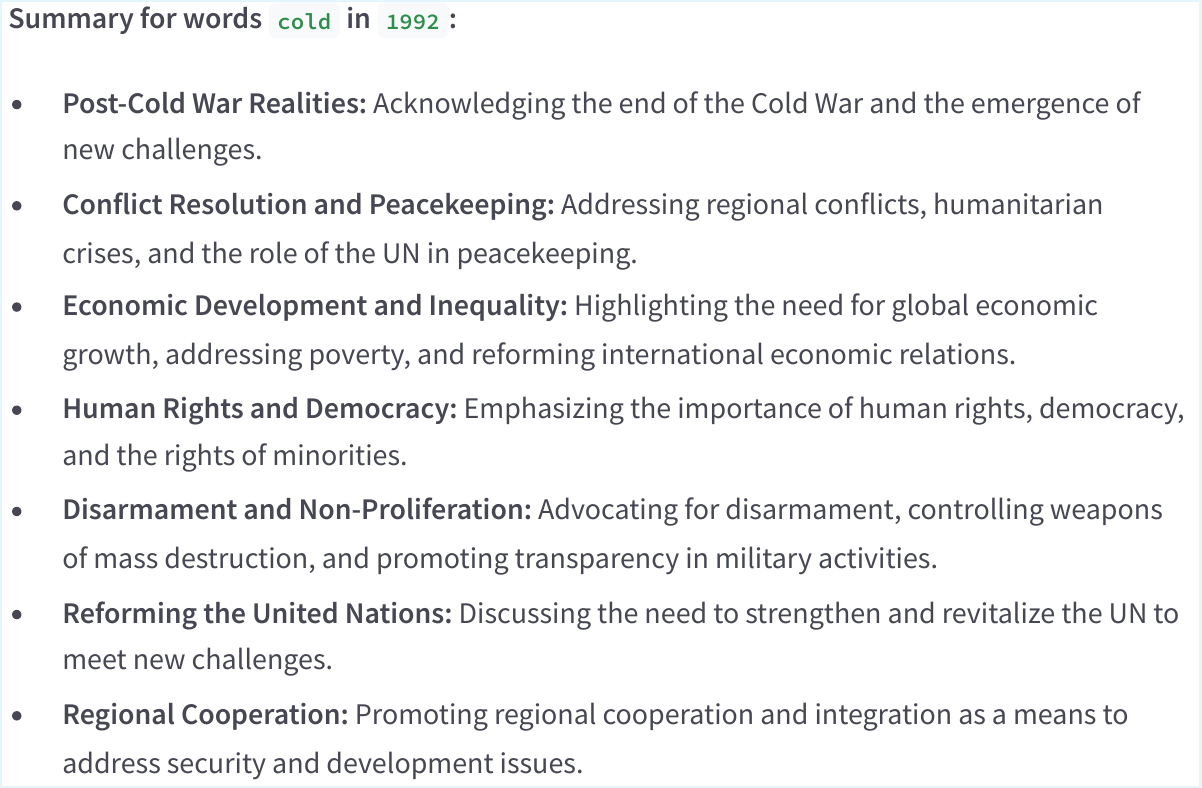}}
    \caption{Summaries generated by DTECT for 1992, centered on post-Cold War discourse.}
    \label{fig:summary_cold}
\end{figure}

\paragraph{Post-Cold War Reflections and Summary Insights}
In 1992, DTECT highlights the term \textit{``cold"} as central to global political discourse, especially regarding the Cold War’s end. The retrieved summaries emphasize post-Cold War realignment, regional conflicts, peacekeeping, democratization, economic inequality, disarmament, and institutional reform (Figure~\ref{fig:summary_cold}).

\begin{figure}[!ht]
    \centering
    \fbox{\includegraphics[width=\linewidth]{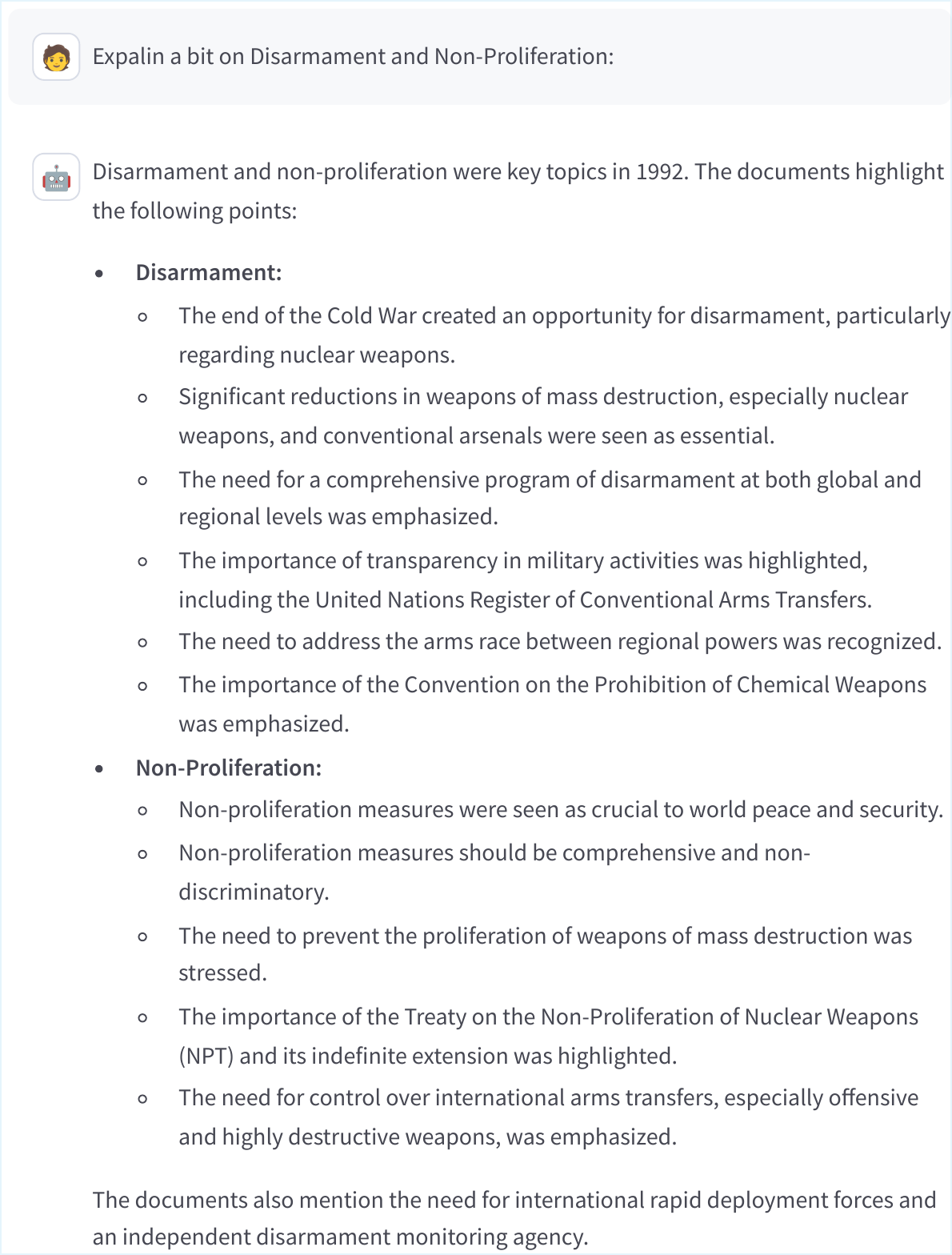}}
    \caption{DTECT’s chat assistant explains post-Cold War disarmament and non-proliferation efforts.}
    \label{fig:followup_cold}
\end{figure}

\paragraph{Follow-up Exploration: Disarmament and Non-Proliferation}
A follow-up query on “Disarmament and Non-Proliferation” reveals a broad set of proposals from the 1992 debates (Figure~\ref{fig:followup_cold}). The end of the Cold War was seen as a key moment to advance disarmament and strengthen arms control frameworks.

\begin{itemize}
    \item \textbf{Disarmament:} Emphasis on reducing nuclear and conventional weapons, promoting transparency (e.g., UN Arms Register), and addressing regional arms races. The Chemical Weapons Convention was widely endorsed.
    
    \item \textbf{Non-Proliferation:} Support for a non-discriminatory approach, indefinite extension of the NPT, and tighter controls on transfers of destructive weapons.
    
    \item \textbf{Institutional Measures:} Calls for rapid deployment forces and an independent disarmament monitoring body to ensure implementation and accountability.
\end{itemize}
These priorities reflect the UN’s evolving role in advancing global peace and security through cooperative disarmament efforts.

\end{document}